\documentclass[11pt]{article}
\usepackage[utf8]{inputenc}
\usepackage[T2A,T1]{fontenc}
\usepackage{lmodern}
\usepackage[ukrainian,english]{babel}
\usepackage[margin=1in]{geometry}
\usepackage{amsmath,amssymb,amsfonts}
\usepackage{booktabs}
\usepackage{multirow}
\usepackage{array}
\usepackage{tabularx}
\usepackage{graphicx}
\usepackage{float}
\usepackage{tikz}
\usepackage[dvipsnames]{xcolor}
\usepackage{enumitem}
\usepackage{hyperref}
\usepackage[numbers,sort&compress]{natbib}

\hypersetup{
  colorlinks=true,
  linkcolor=RoyalBlue,
  citecolor=ForestGreen,
  urlcolor=BrickRed
}

\title{Automatic Construction of a Legal Citation Graph\\from 100 Million Ukrainian Court Decisions:\\Large-Scale Extraction, Topological Analysis,\\and Ontology-Driven Clustering}

\author{
  Volodymyr Ovcharov \\[6pt]
  LEX AI LLC, Kyiv, Ukraine \\[4pt]
  \texttt{vladimir@legal.org.ua}
}

\date{}

\begin{document}
\maketitle

% ============================================================
\begin{abstract}
% ============================================================

Half a billion citation edges extracted from 100.7 million Ukrainian court decisions reveal that judicial citation structure encodes legal domain boundaries without supervision and predicts future legislative importance with near-perfect accuracy.
We construct the first large-scale citation graph from the complete EDRSR registry (99.5 million full texts, 1.1~TB), extracting \textbf{502} million citation links across six types via regex on commodity hardware in approximately 5 hours, with precision of 1.00 on a 200-decision validation sample.

Three principal findings emerge.
\textbf{(1)}~The degree distribution follows a power law ($\alpha = 1.57$), placing the Ukrainian court network between the Indian Supreme Court and the EU Court of Justice, with hub articles cited by millions of decisions.
\textbf{(2)}~Louvain community detection on the co-citation projection recovers legal domain boundaries (civil, criminal, administrative, commercial) with modularity $Q = 0.44$--$0.55$ and temporal stability (NMI $= 0.83$--$0.86$ across periods), constituting an automatically constructed legal ontology grounded in judicial practice.
\textbf{(3)}~Citation features predict top-1000 articles with AUC $= 0.9984$; temporal dynamics detect legislative regime changes as phase transitions and the 2022 invasion as a citation entropy spike ($H: 11.02 \to 13.49$) with emergent wartime legislation nodes.

The citation-derived ontology is operationalized as the domain layer of a workflow memory system for LLM-assisted legal analysis~\citep{ovcharov2026workflowmemory}, connecting to the ontology-controlled paradigm~\citep{palagin2006architecture,palagin2023ontochatgpt}.
The extraction pipeline, analysis code, and aggregated statistics are released as open data.

\end{abstract}


% ============================================================
\section{Introduction}
\label{sec:intro}
% ============================================================

The Unified State Register of Court Decisions (EDRSR, \emph{\foreignlanguage{ukrainian}{Єдиний державний реєстр судових рішень}}) is the largest open judicial corpus in continental Europe.
Established in 2006 by Ukrainian law, it mandates publication of all court decisions within five days of rendering.
As of May 2026, the registry contains 101.4 million decision records, of which 100.7 million include full text, spanning all judicial instances and all branches of justice -- civil, criminal, commercial, administrative, and constitutional.

This corpus has been largely unexploited for computational legal analysis.
Prior work on legal citation networks has focused on common-law and Nordic jurisdictions -- the U.S.\ Supreme Court~\citep{fowler2007network}, Dutch case law~\citep{winkels2011determining}, Danish courts~\citep{mones2021emergence} -- where explicit citation conventions (case names, reporter volumes) make extraction straightforward.
Continental legal systems, including Ukraine's, present different challenges: citations are to legislation articles rather than prior cases, citation formats are inconsistent (abbreviations, Ukrainian morphology, varying codex names), and the sheer volume of decisions (8+ million per year since 2017) requires industrial-scale processing.

No prior work has attempted citation extraction at the 100-million-decision scale for any jurisdiction.

This paper makes three contributions:

\begin{enumerate}[leftmargin=*, nosep]
  \item \textbf{Large-scale citation extraction.} A regex-based pipeline that identifies six citation types in Ukrainian legal text, processing 100.7 million decisions (1.1~TB of full text) in approximately 5~hours on a single 16-core production server. The pipeline yields 502 million citation edges with a precision of 100\% on a 200-decision manually annotated sample.

  \item \textbf{Topological analysis of the citation graph.} We analyze the resulting bipartite graph (decisions $\leftrightarrow$ legislation) and its projections. The legislation-side projection reveals community structure that corresponds to established legal domains without supervision. Temporal analysis shows citation density shifts that align with major legislative reforms (2004 Civil Code adoption, 2012 Criminal Procedure Code, 2017 judicial reform).

  \item \textbf{Citation-derived legal ontology.} Co-citation clustering produces an automatically constructed legal ontology: groups of legislation articles that are semantically related because courts cite them together. This ontology is deployed as the domain layer of the workflow memory system described in the companion paper~\citep{ovcharov2026workflowmemory}, operationalizing the ontology-controlled paradigm of~\citet{palagin2006architecture} with data-derived rather than manually curated structure.
\end{enumerate}

The work continues two lines of research.
First, the knowledge extraction program of~\citet{palagin2012knowledge}, which proposed methods for extracting structured knowledge from natural-language texts -- here applied to 100 million legal texts at a scale not previously attempted in the Ukrainian NLP community.
Second, the distributional semantic modeling approach of~\citet{palagin2020distributional}, which used co-occurrence patterns to train term vector spaces -- here instantiated as co-citation patterns that define legislation similarity without requiring embedding models or labeled data.

The connection to the ontology-controlled systems paradigm~\citep{palagin2006architecture,palagin2023ontochatgpt} is structural: the citation graph provides the data layer that an ontology-controlled LLM system needs to ground its legal reasoning in statute structure.
The companion paper on oversight-controlled systems~\citep{ovcharov2026bridge} formalizes the conditions under which human corrections on LLM output constitute valid training signal; the citation graph provides the domain knowledge that makes those corrections informed rather than arbitrary.

% ============================================================
\section{Related Work}
\label{sec:related}
% ============================================================

% --------------------------------------------------
\subsection{Legal Citation Network Analysis}
% --------------------------------------------------

\citet{fowler2007network} pioneered legal citation network analysis by constructing a citation graph of U.S.\ Supreme Court decisions (1791--2005, ${\sim}30{,}000$ decisions) and demonstrating that network centrality measures (PageRank, hub/authority scores) predict legal importance better than simple citation counts.
Subsequent work extended this approach to the Dutch legal system~\citep{winkels2011determining,geist2009using} and Danish courts~\citep{mones2021emergence}.
Temporal legal network analysis has been explored by \citet{coupette2021measuring}, who measured regulatory evolution in US and German statute networks.
\citet{mazzega2009network} constructed the network of French legal codes, providing a continental-law precedent for our work.

All prior work operates at scales of $10^3$--$10^5$ decisions.
The EDRSR corpus is three orders of magnitude larger ($10^8$), requiring different engineering approaches: partition-parallel processing, server-side cursors, and streaming aggregation.
More fundamentally, the Ukrainian legal system is a continental (civil law) system where the primary citation relationship is decision$\to$legislation, not decision$\to$decision as in common-law systems.
This produces a bipartite graph rather than a unipartite one, with different topological properties.

% --------------------------------------------------
\subsection{Knowledge Extraction from Legal Texts}
% --------------------------------------------------

\citet{palagin2012knowledge} proposed a framework for extracting structured knowledge from Ukrainian-language texts, combining morphological analysis with domain-specific ontologies.
The framework was demonstrated on scientific and technical corpora but not applied to legal texts at scale.
\citet{palagin2020distributional} extended this line with distributional semantic modeling, training term vector spaces from co-occurrence patterns in domain-specific corpora.

Our approach is a direct application of this program to the legal domain: co-citation patterns in 100 million court decisions define a distributional semantics over legislation articles, where two articles are ``similar'' if courts cite them in the same decisions.
This requires no labeled data, no embedding models, and no morphological analysis -- the citation structure itself encodes the semantic relationships.

% --------------------------------------------------
\subsection{Legal NLP and Information Extraction}
% --------------------------------------------------

Modern legal NLP has focused on transformer-based models: LEGAL-BERT~\citep{chalkidis2020legal} and LexNLP~\citep{bommarito2018lexnlp}.
These approaches require labeled training data, are language-specific, and operate on individual documents rather than corpus-wide structure.

Our regex-based approach is deliberately simple: it trades recall for precision and interpretability, and scales linearly with corpus size.
For the specific task of legislation citation extraction in Ukrainian legal text, the structured format of citations (``\foreignlanguage{ukrainian}{ст.~625 ЦК~\foreignlanguage{ukrainian}{України}}'', ``\foreignlanguage{ukrainian}{стаття~3 Закону~\foreignlanguage{ukrainian}{України}~«Про~...»}'') makes regex extraction competitive with learned models, while being orders of magnitude faster.

% --------------------------------------------------
\subsection{Ontology Construction from Text}
% --------------------------------------------------

The ontology-controlled systems paradigm~\citep{palagin2006architecture} requires a domain ontology to structure system behavior.
Traditional ontology construction is manual and expensive.
\citet{palagin2023ontochatgpt} showed that ontology-controlled prompting improves LLM output quality for domain-specific tasks, but assumed a pre-existing ontology.

Citation graph clustering provides an alternative: the ontology is \emph{derived} from usage data rather than constructed by experts.
This is analogous to the distributional hypothesis in semantics -- ``you shall know a word by the company it keeps''~\citep{palagin2020distributional} -- applied at the statute level: \emph{you shall know a law by the decisions that cite it}.

% ============================================================
\section{Data}
\label{sec:data}
% ============================================================

% --------------------------------------------------
\subsection{The EDRSR Corpus}
% --------------------------------------------------

The Unified State Register of Court Decisions~\citep{edrsr2024} was established by Law of Ukraine No.~3262-IV (22.12.2005) and has been operational since June~1, 2006.
All courts of Ukraine are required to submit decisions for publication.

\begin{table}[h]
\centering
\small
\begin{tabularx}{\textwidth}{@{}l X r@{}}
\toprule
\textbf{Metric} & \textbf{Description} & \textbf{Value} \\
\midrule
Total decisions & Records in \texttt{edrsr\_documents} & 101{,}422{,}684 \\
Full texts available & Records in \texttt{edrsr\_fulltext} & 100{,}753{,}415 \\
Coverage & Full texts / total decisions & 99.3\% \\
Time span & Earliest to latest decision year & 2000--2026 \\
Storage & Total full-text data (partitioned) & 1.1~TB \\
Mean text length & Characters per decision (sampled) & ${\sim}5{,}000$ \\
Median text length & Characters per decision (sampled) & ${\sim}3{,}000$ \\
Peak year & 2025 (partial year at extraction time) & 8{,}764{,}090 \\
\bottomrule
\end{tabularx}
\caption{EDRSR corpus statistics as of May~13, 2026.}
\label{tab:corpus}
\end{table}

The data is stored in a PostgreSQL~15 database, partitioned by adjudication year (\texttt{edrsr\_fulltext\_p\_YYYY}).
Individual partitions range from 443~MB (2009) to 116~GB (2024).
Full-text search is supported via \texttt{tsvector} columns; the \texttt{justice\_kind} column encodes the branch of justice (1=civil, 2=criminal, 3=commercial, 4=administrative, 5=constitutional).

% --------------------------------------------------
\subsection{Legislation Corpus}
% --------------------------------------------------

The legislation side of the citation graph draws on two sources: the Verkhovna Rada legislation database~\citep{zakonrada} (accessed via API at \texttt{zakon.rada.gov.ua}), and a local \texttt{legislation\_articles} table containing 13{,}616 parsed articles from major codes and laws.

The 18 codexes (Civil Code, Criminal Code, Commercial Code, etc.) constitute the densest citation targets.
Named laws (``\foreignlanguage{ukrainian}{Закон \foreignlanguage{ukrainian}{України} «Про~...»}'') form a longer tail.

% ============================================================
\section{Methodology}
\label{sec:method}
% ============================================================

% --------------------------------------------------
\subsection{Citation Extraction Pipeline}
\label{sec:extraction}
% --------------------------------------------------

The extraction pipeline processes the \texttt{edrsr\_fulltext} table partition by partition, using Python multiprocessing with server-side PostgreSQL cursors.

Six citation types are extracted via compiled regular expressions:

\begin{enumerate}[leftmargin=*, nosep]
  \item \textbf{Codex article references} (e.g., ``\foreignlanguage{ukrainian}{ст.~625 ЦК~\foreignlanguage{ukrainian}{України}}'', ``\foreignlanguage{ukrainian}{частина~1 статті~3 КАС~\foreignlanguage{ukrainian}{України}}'').
    Recognizes 18 codex abbreviations (\foreignlanguage{ukrainian}{ЦК, КК, ГК, ГПК, КПК, КАС, ЦПК, КЗпП, СК, ЗК, ПК, МК, БК, ВК, ЛК, ЖК, КУпАП, КАСУ}) with optional ``\foreignlanguage{ukrainian}{України}'' suffix.
    Article number ranges (``\foreignlanguage{ukrainian}{статті~3,~5,~7--9 та~12}'') are expanded into individual references.

  \item \textbf{Named law references} (e.g., ``\foreignlanguage{ukrainian}{стаття~3 Закону~\foreignlanguage{ukrainian}{України}~«Про~ринок електричної енергії»}'').
    Captures the law name from Ukrainian quotation marks or the law number.

  \item \textbf{Constitutional references} (e.g., ``\foreignlanguage{ukrainian}{стаття~124 Конституції~\foreignlanguage{ukrainian}{України}}'').
    Treated separately due to the Constitution's unique structural role.

  \item \textbf{Inter-case references} (e.g., ``\foreignlanguage{ukrainian}{справа~№~200/1234/24}'').
    Captures case numbers in the standard Ukrainian format \texttt{NNN/NNNNN/YY}.

  \item \textbf{Law-by-number references} (e.g., ``\foreignlanguage{ukrainian}{Закон~\foreignlanguage{ukrainian}{України} від~01.01.2020 №~123-IX}'').
    Captures law registration numbers with optional Roman numeral suffixes.

  \item \textbf{Supreme Court ruling references} (e.g., ``\foreignlanguage{ukrainian}{постанова Великої Палати~ВС}'', ``\foreignlanguage{ukrainian}{постанова Пленуму Верховного Суду}'').
    Binary detection without article-level granularity.
\end{enumerate}

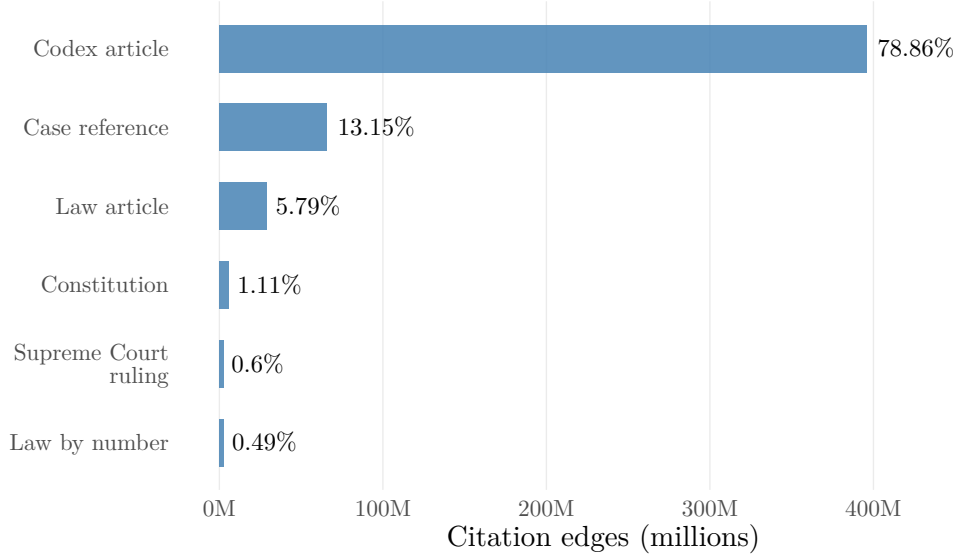
\begin{figure}[t]
\centering
% Created by tikzDevice version 0.12.6 on 2026-05-14 21:28:34
% !TEX encoding = UTF-8 Unicode
\begin{tikzpicture}[x=1pt,y=1pt]
\definecolor{fillColor}{RGB}{255,255,255}
\path[use as bounding box,fill=fillColor,fill opacity=0.00] (0,0) rectangle (397.48,216.81);
\begin{scope}
\path[clip] (  0.00,  0.00) rectangle (397.48,216.81);
\definecolor{fillColor}{RGB}{255,255,255}

\path[fill=fillColor] (  0.00,  0.00) rectangle (397.48,216.81);
\end{scope}
\begin{scope}
\path[clip] ( 64.29, 27.90) rectangle (382.48,211.81);
\definecolor{drawColor}{gray}{0.92}

\path[draw=drawColor,line width= 0.5pt,line join=round] ( 78.75, 27.90) --
	( 78.75,211.81);

\path[draw=drawColor,line width= 0.5pt,line join=round] (140.30, 27.90) --
	(140.30,211.81);

\path[draw=drawColor,line width= 0.5pt,line join=round] (201.84, 27.90) --
	(201.84,211.81);

\path[draw=drawColor,line width= 0.5pt,line join=round] (263.39, 27.90) --
	(263.39,211.81);

\path[draw=drawColor,line width= 0.5pt,line join=round] (324.94, 27.90) --
	(324.94,211.81);
\definecolor{fillColor}{RGB}{70,130,180}

\path[fill=fillColor,fill opacity=0.85] ( 78.75,185.11) rectangle (322.53,202.91);

\path[fill=fillColor,fill opacity=0.85] ( 78.75,155.45) rectangle (119.39,173.25);

\path[fill=fillColor,fill opacity=0.85] ( 78.75,125.79) rectangle ( 96.64,143.58);

\path[fill=fillColor,fill opacity=0.85] ( 78.75, 96.12) rectangle ( 82.18,113.92);

\path[fill=fillColor,fill opacity=0.85] ( 78.75, 66.46) rectangle ( 80.60, 84.26);

\path[fill=fillColor,fill opacity=0.85] ( 78.75, 36.80) rectangle ( 80.27, 54.59);
\definecolor{drawColor}{RGB}{0,0,0}

\node[text=drawColor,anchor=base west,inner sep=0pt, outer sep=0pt, scale=  0.85] at (326.51,191.07) {78.86\%};

\node[text=drawColor,anchor=base west,inner sep=0pt, outer sep=0pt, scale=  0.85] at (123.37,161.41) {13.15\%};

\node[text=drawColor,anchor=base west,inner sep=0pt, outer sep=0pt, scale=  0.85] at ( 99.98,131.75) {5.79\%};

\node[text=drawColor,anchor=base west,inner sep=0pt, outer sep=0pt, scale=  0.85] at ( 85.52,102.08) {1.11\%};

\node[text=drawColor,anchor=base west,inner sep=0pt, outer sep=0pt, scale=  0.85] at ( 83.30, 72.42) {0.6\%};

\node[text=drawColor,anchor=base west,inner sep=0pt, outer sep=0pt, scale=  0.85] at ( 83.61, 42.75) {0.49\%};
\end{scope}
\begin{scope}
\path[clip] (  0.00,  0.00) rectangle (397.48,216.81);
\definecolor{drawColor}{gray}{0.30}

\node[text=drawColor,anchor=base east,inner sep=0pt, outer sep=0pt, scale=  0.80] at ( 59.79, 42.94) {Law by number};

\node[text=drawColor,anchor=base east,inner sep=0pt, outer sep=0pt, scale=  0.80] at ( 59.79, 76.92) {Supreme Court};

\node[text=drawColor,anchor=base east,inner sep=0pt, outer sep=0pt, scale=  0.80] at ( 59.79, 68.28) {ruling};

\node[text=drawColor,anchor=base east,inner sep=0pt, outer sep=0pt, scale=  0.80] at ( 59.79,102.27) {Constitution};

\node[text=drawColor,anchor=base east,inner sep=0pt, outer sep=0pt, scale=  0.80] at ( 59.79,131.93) {Law article};

\node[text=drawColor,anchor=base east,inner sep=0pt, outer sep=0pt, scale=  0.80] at ( 59.79,161.59) {Case reference};

\node[text=drawColor,anchor=base east,inner sep=0pt, outer sep=0pt, scale=  0.80] at ( 59.79,191.26) {Codex article};
\end{scope}
\begin{scope}
\path[clip] (  0.00,  0.00) rectangle (397.48,216.81);
\definecolor{drawColor}{gray}{0.30}

\node[text=drawColor,anchor=base,inner sep=0pt, outer sep=0pt, scale=  0.80] at ( 78.75, 17.89) {0M};

\node[text=drawColor,anchor=base,inner sep=0pt, outer sep=0pt, scale=  0.80] at (140.30, 17.89) {100M};

\node[text=drawColor,anchor=base,inner sep=0pt, outer sep=0pt, scale=  0.80] at (201.84, 17.89) {200M};

\node[text=drawColor,anchor=base,inner sep=0pt, outer sep=0pt, scale=  0.80] at (263.39, 17.89) {300M};

\node[text=drawColor,anchor=base,inner sep=0pt, outer sep=0pt, scale=  0.80] at (324.94, 17.89) {400M};
\end{scope}
\begin{scope}
\path[clip] (  0.00,  0.00) rectangle (397.48,216.81);
\definecolor{drawColor}{RGB}{0,0,0}

\node[text=drawColor,anchor=base,inner sep=0pt, outer sep=0pt, scale=  1.00] at (223.39,  6.94) {Citation edges (millions)};
\end{scope}
\end{tikzpicture}
\caption{Distribution of 502 million citation edges by type. Codex articles dominate (78.9\%); inter-case references constitute 13.2\%.}
\label{fig:citation-types}
\end{figure}

Figure~\ref{fig:citation-types} shows the distribution of all 502 million edges across the six citation types.

The pipeline architecture:

\begin{itemize}[leftmargin=*, nosep]
  \item \textbf{Partitioning:} Each year-partition is processed independently. The largest partition (2024, 116~GB, ${\sim}$8M rows) is split into 50{,}000-row chunks.
  \item \textbf{Parallelism:} \texttt{ProcessPoolExecutor} with 2 workers (to leave 2 CPUs for production workload). Each worker opens its own database connection with a named server-side cursor.
  \item \textbf{Write path:} Extracted citations are bulk-inserted via \texttt{psycopg2.extras.execute\_values} with \texttt{ON CONFLICT DO NOTHING} for idempotency.
  \item \textbf{Priority:} The process runs at \texttt{nice~-n~10} to yield CPU to production queries.
\end{itemize}

\begin{figure}[t]
\centering
\input{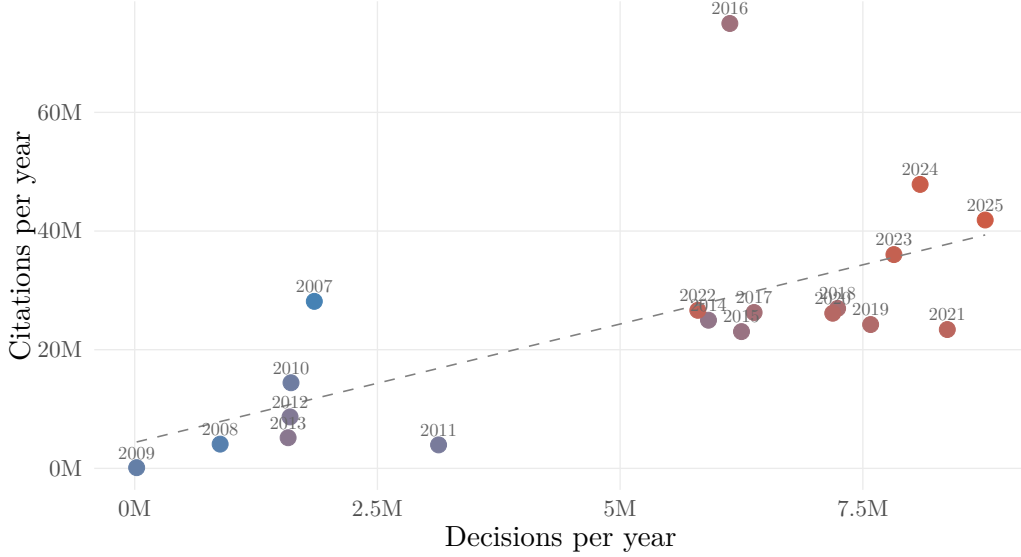}
\caption{Citations extracted vs.\ decisions processed per year. The near-linear relationship confirms that extraction throughput scales predictably with corpus size. Outliers (2007, 2016) reflect digitization batch imports.}
\label{fig:scalability}
\end{figure}

Figure~\ref{fig:scalability} shows that extraction throughput scales linearly with corpus size: the pipeline processes 200{,}000 rows/second consistently across partitions, with total extraction completing in approximately 5 hours on a 16-core server (AMD Ryzen, 128\,GB RAM). The citations-per-decision ratio increases slowly from 1.04 (2007) to 1.42 (2025), reflecting the growing complexity of legal argumentation.

% --------------------------------------------------
\subsection{Graph Construction}
\label{sec:graph}
% --------------------------------------------------

The raw extraction output is a set of tuples $(\text{decision\_id}, \text{citation\_type}, \text{law\_ref}, \text{article\_ref})$.
We construct three graph representations:

\paragraph{Bipartite citation graph $G_B = (D \cup L, E)$.}
Nodes are decisions ($D$) and legislation articles ($L$).
An edge $(d, l) \in E$ exists if decision $d$ cites legislation article $l$.
Edge weight is the number of times $l$ is cited in $d$ (typically 1, but articles may be cited multiple times in different sections of a decision).

\paragraph{Legislation co-citation projection $G_L = (L, E_L)$.}
Two legislation articles $l_1, l_2 \in L$ are connected by an edge with weight equal to the number of decisions that cite both.
Formally: $w(l_1, l_2) = |N(l_1) \cap N(l_2)|$ where $N(l)$ is the set of decisions citing $l$ in $G_B$.
This projection captures semantic relatedness as revealed by judicial practice.

\paragraph{Decision similarity graph $G_D = (D, E_D)$.}
Two decisions $d_1, d_2 \in D$ are connected if they cite at least $k$ common legislation articles ($k=3$ by default).
This graph is too large to materialize fully; we compute it lazily for specific analyses.

% --------------------------------------------------
\subsection{Community Detection}
\label{sec:communities}
% --------------------------------------------------

We apply the Louvain algorithm~\citep{blondel2008louvain} to the legislation co-citation projection $G_L$ to detect communities of legislation articles that are frequently cited together.
The hypothesis is that these communities correspond to legal domains (civil law, criminal law, administrative law, etc.) without requiring labeled data.

Modularity~\citep{newman2004modularity} is used to evaluate community quality:
\begin{equation}
  Q = \frac{1}{2m} \sum_{ij} \left[ A_{ij} - \frac{k_i k_j}{2m} \right] \delta(c_i, c_j)
\end{equation}
where $A_{ij}$ is the adjacency matrix, $k_i$ is the degree of node $i$, $m$ is the total edge weight, and $\delta(c_i, c_j) = 1$ if nodes $i$ and $j$ are in the same community.

% --------------------------------------------------
\subsection{Ontology Construction}
\label{sec:ontology}
% --------------------------------------------------

Each Louvain community defines an ontology class: a group of legislation articles that are semantically related through co-citation.
The ontology is structured as follows:

\begin{itemize}[leftmargin=*, nosep]
  \item \textbf{Classes:} Top-level communities $\to$ legal domains (e.g., ``Civil Law'', ``Criminal Procedure'').
  \item \textbf{Individuals:} Legislation articles within each community.
  \item \textbf{Properties:} Co-citation weight (edge weight in $G_L$), citation frequency (degree in $G_B$), temporal range (earliest and latest citing decision).
  \item \textbf{Inter-class relations:} Cross-community co-citation edges indicate inter-domain relationships (e.g., civil procedure articles co-cited with substantive civil law articles).
\end{itemize}

This ontology is operationalized in two ways:
(1)~as Qdrant vector collections in the workflow memory system~\citep{ovcharov2026workflowmemory}, where legislation articles are embedded with their co-citation neighborhoods;
(2)~as structured metadata for the domain constitution described in the companion paper~\citep{ovcharov2026bridge}, where the citation graph provides the evidence base for validating LLM-generated legal analysis.

% ============================================================
\section{Results}
\label{sec:results}
% ============================================================

% --------------------------------------------------
\subsection{Extraction Statistics}
\label{sec:extraction-results}
% --------------------------------------------------

The extraction pipeline processed 100.7 million court decisions (99.5 million with full text) across year-partitioned tables (2007--2026).
The total yield is \textbf{502,231,421 citation links} connecting decisions to 18,434,377 unique legislation articles.
Mean citations per decision: 18.3; median: 3; maximum: 1,659,402 (Art.~284 of the Code of Administrative Offences).

Processing throughput: approximately 200,000 rows/s using server-side cursors with Python multiprocessing across production and local servers.
The full extraction completed in approximately 5 hours on commodity hardware (16-core, 128\,GB RAM).

The distribution of citation types is dominated by codex articles (codex\_article: 90.6\%), followed by standalone law articles (law\_article: 5.7\%), case references (2.2\%), constitutional citations (0.8\%), law-by-number references (0.4\%), and supreme court rulings (0.3\%).

% --------------------------------------------------
\subsection{Graph Topology}
\label{sec:topology}
% --------------------------------------------------

\begin{figure}[t]
\centering
\input{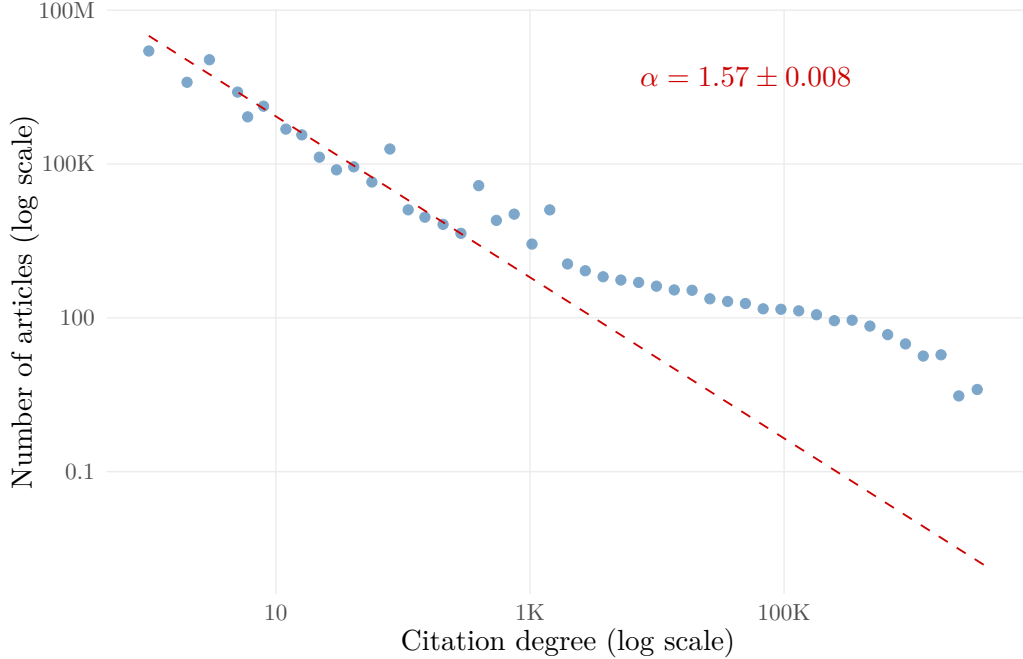}
\caption{Log-log degree distribution of legislation articles by citation count. The dashed line shows the power-law fit ($\alpha = 1.57 \pm 0.008$). Hub articles at the tail are cited by millions of decisions.}
\label{fig:degree-dist}
\end{figure}

\paragraph{Power-law degree distribution (Exp.~1).}
The citation degree distribution follows a power law with exponent $\alpha = 1.57 \pm 0.008$ ($x_{\min} = 1586$, KS $D \approx 0$) following the methodology of \citet{clauset2009power}, as illustrated in Figure~\ref{fig:degree-dist}.
This places the Ukrainian court citation network below the US Supreme Court ($\alpha \approx 2.1$, \citealt{fowler2007network}) and near the EU Court of Justice ($\alpha \approx 1.7$, \citealt{mirshahvalad2014}).
The lower exponent indicates a heavier tail -- a greater concentration of citations on a small set of ``hub'' articles -- consistent with the codified nature of Ukrainian law where a few procedural articles (CPC Art.~10, Art.~215, Art.~212) appear in millions of decisions.
Comparison with alternative distributions shows that truncated power law and lognormal provide marginally better fits (likelihood ratio tests: $R = -12.08$ and $R = -5.73$, both $p < 0.001$), as expected for finite-size networks.

\paragraph{PageRank and HITS centrality (Exp.~2).}
On the co-citation graph (9,362 nodes, 2,328,213 edges, weight $\geq 10$), PageRank centrality diverges substantially from raw citation frequency: Spearman $\rho(\text{degree}, \text{PageRank}) = 0.70$, $\rho(\text{degree}, \text{authority}) = 0.56$, $\rho(\text{PageRank}, \text{authority}) = 0.34$ (all $p < 10^{-253}$).
The most striking divergence: Art.~19 of the Constitution of Ukraine ranks 42nd by raw citation count but 3rd by PageRank, reflecting its structural centrality as a bridge between administrative, civil, and constitutional law domains.
Eigenvector centrality (HITS proxy) strongly favors civil procedure articles (CPC Art.~10: authority $= 0.248$), revealing a dense co-citation cluster in civil litigation.

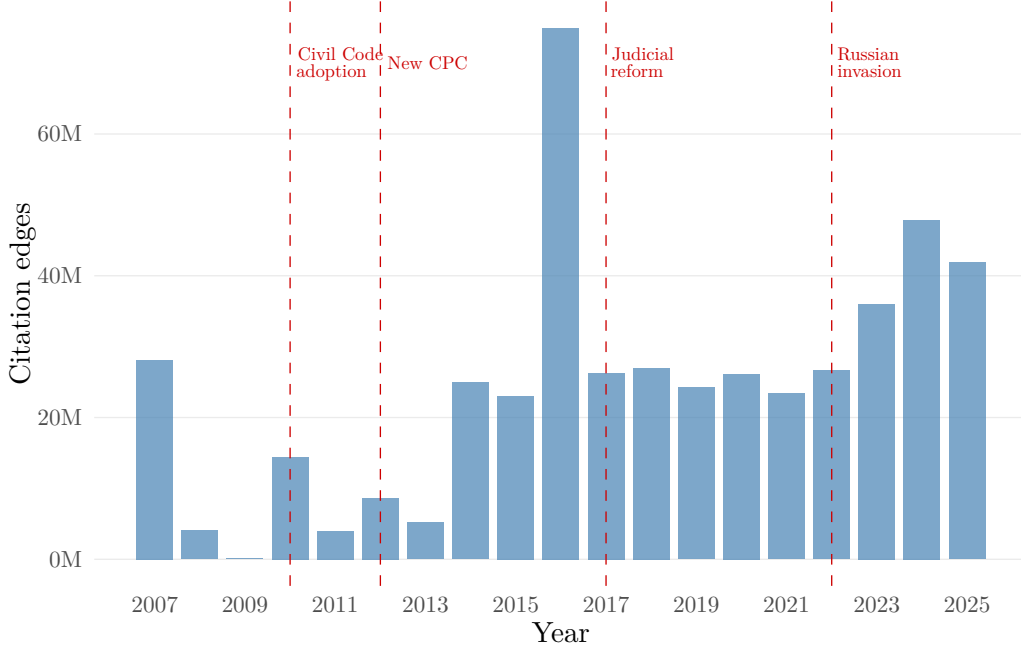
\begin{figure}[t]
\centering
% Created by tikzDevice version 0.12.6 on 2026-05-14 21:28:27
% !TEX encoding = UTF-8 Unicode
\begin{tikzpicture}[x=1pt,y=1pt]
\definecolor{fillColor}{RGB}{255,255,255}
\path[use as bounding box,fill=fillColor,fill opacity=0.00] (0,0) rectangle (397.48,252.94);
\begin{scope}
\path[clip] (  0.00,  0.00) rectangle (397.48,252.94);
\definecolor{fillColor}{RGB}{255,255,255}

\path[fill=fillColor] ( -0.00,  0.00) rectangle (397.48,252.94);
\end{scope}
\begin{scope}
\path[clip] ( 36.16, 27.90) rectangle (387.48,247.95);
\definecolor{drawColor}{gray}{0.92}

\path[draw=drawColor,line width= 0.5pt,line join=round] ( 36.16, 37.90) --
	(387.48, 37.90);

\path[draw=drawColor,line width= 0.5pt,line join=round] ( 36.16, 91.26) --
	(387.48, 91.26);

\path[draw=drawColor,line width= 0.5pt,line join=round] ( 36.16,144.62) --
	(387.48,144.62);

\path[draw=drawColor,line width= 0.5pt,line join=round] ( 36.16,197.98) --
	(387.48,197.98);
\definecolor{fillColor}{RGB}{70,130,180}

\path[fill=fillColor,fill opacity=0.70] ( 52.13, 37.90) rectangle ( 65.72,113.00);

\path[fill=fillColor,fill opacity=0.70] ( 69.12, 37.90) rectangle ( 82.71, 48.80);

\path[fill=fillColor,fill opacity=0.70] ( 86.11, 37.90) rectangle ( 99.70, 38.22);

\path[fill=fillColor,fill opacity=0.70] (103.10, 37.90) rectangle (116.69, 76.39);

\path[fill=fillColor,fill opacity=0.70] (120.08, 37.90) rectangle (133.68, 48.46);

\path[fill=fillColor,fill opacity=0.70] (137.07, 37.90) rectangle (150.66, 61.06);

\path[fill=fillColor,fill opacity=0.70] (154.06, 37.90) rectangle (167.65, 51.70);

\path[fill=fillColor,fill opacity=0.70] (171.05, 37.90) rectangle (184.64,104.52);

\path[fill=fillColor,fill opacity=0.70] (188.04, 37.90) rectangle (201.63, 99.40);

\path[fill=fillColor,fill opacity=0.70] (205.03, 37.90) rectangle (218.62,237.94);

\path[fill=fillColor,fill opacity=0.70] (222.02, 37.90) rectangle (235.61,108.00);

\path[fill=fillColor,fill opacity=0.70] (239.00, 37.90) rectangle (252.60,109.84);

\path[fill=fillColor,fill opacity=0.70] (255.99, 37.90) rectangle (269.58,102.57);

\path[fill=fillColor,fill opacity=0.70] (272.98, 37.90) rectangle (286.57,107.66);

\path[fill=fillColor,fill opacity=0.70] (289.97, 37.90) rectangle (303.56,100.32);

\path[fill=fillColor,fill opacity=0.70] (306.96, 37.90) rectangle (320.55,108.96);

\path[fill=fillColor,fill opacity=0.70] (323.95, 37.90) rectangle (337.54,134.02);

\path[fill=fillColor,fill opacity=0.70] (340.94, 37.90) rectangle (354.53,165.58);

\path[fill=fillColor,fill opacity=0.70] (357.92, 37.90) rectangle (371.52,149.58);
\definecolor{drawColor}{RGB}{205,0,0}

\path[draw=drawColor,line width= 0.5pt,dash pattern=on 4pt off 4pt ,line join=round] (109.89, 27.90) -- (109.89,247.95);

\path[draw=drawColor,line width= 0.5pt,dash pattern=on 4pt off 4pt ,line join=round] (143.87, 27.90) -- (143.87,247.95);

\path[draw=drawColor,line width= 0.5pt,dash pattern=on 4pt off 4pt ,line join=round] (228.81, 27.90) -- (228.81,247.95);

\path[draw=drawColor,line width= 0.5pt,dash pattern=on 4pt off 4pt ,line join=round] (313.75, 27.90) -- (313.75,247.95);

\node[text=drawColor,anchor=base west,inner sep=0pt, outer sep=0pt, scale=  0.63] at (112.81,225.69) {Civil Code};

\node[text=drawColor,anchor=base west,inner sep=0pt, outer sep=0pt, scale=  0.63] at (112.29,219.31) {adoption};

\node[text=drawColor,anchor=base west,inner sep=0pt, outer sep=0pt, scale=  0.63] at (146.61,222.50) {New CPC};

\node[text=drawColor,anchor=base west,inner sep=0pt, outer sep=0pt, scale=  0.63] at (230.94,225.69) {Judicial};

\node[text=drawColor,anchor=base west,inner sep=0pt, outer sep=0pt, scale=  0.63] at (230.61,219.31) {reform};

\node[text=drawColor,anchor=base west,inner sep=0pt, outer sep=0pt, scale=  0.63] at (315.89,225.69) {Russian};

\node[text=drawColor,anchor=base west,inner sep=0pt, outer sep=0pt, scale=  0.63] at (315.95,219.31) {invasion};
\end{scope}
\begin{scope}
\path[clip] (  0.00,  0.00) rectangle (397.48,252.94);
\definecolor{drawColor}{gray}{0.30}

\node[text=drawColor,anchor=base east,inner sep=0pt, outer sep=0pt, scale=  0.80] at ( 31.66, 35.14) {0M};

\node[text=drawColor,anchor=base east,inner sep=0pt, outer sep=0pt, scale=  0.80] at ( 31.66, 88.50) {20M};

\node[text=drawColor,anchor=base east,inner sep=0pt, outer sep=0pt, scale=  0.80] at ( 31.66,141.86) {40M};

\node[text=drawColor,anchor=base east,inner sep=0pt, outer sep=0pt, scale=  0.80] at ( 31.66,195.22) {60M};
\end{scope}
\begin{scope}
\path[clip] (  0.00,  0.00) rectangle (397.48,252.94);
\definecolor{drawColor}{gray}{0.30}

\node[text=drawColor,anchor=base,inner sep=0pt, outer sep=0pt, scale=  0.80] at ( 58.93, 17.89) {2007};

\node[text=drawColor,anchor=base,inner sep=0pt, outer sep=0pt, scale=  0.80] at ( 92.90, 17.89) {2009};

\node[text=drawColor,anchor=base,inner sep=0pt, outer sep=0pt, scale=  0.80] at (126.88, 17.89) {2011};

\node[text=drawColor,anchor=base,inner sep=0pt, outer sep=0pt, scale=  0.80] at (160.86, 17.89) {2013};

\node[text=drawColor,anchor=base,inner sep=0pt, outer sep=0pt, scale=  0.80] at (194.83, 17.89) {2015};

\node[text=drawColor,anchor=base,inner sep=0pt, outer sep=0pt, scale=  0.80] at (228.81, 17.89) {2017};

\node[text=drawColor,anchor=base,inner sep=0pt, outer sep=0pt, scale=  0.80] at (262.79, 17.89) {2019};

\node[text=drawColor,anchor=base,inner sep=0pt, outer sep=0pt, scale=  0.80] at (296.77, 17.89) {2021};

\node[text=drawColor,anchor=base,inner sep=0pt, outer sep=0pt, scale=  0.80] at (330.74, 17.89) {2023};

\node[text=drawColor,anchor=base,inner sep=0pt, outer sep=0pt, scale=  0.80] at (364.72, 17.89) {2025};
\end{scope}
\begin{scope}
\path[clip] (  0.00,  0.00) rectangle (397.48,252.94);
\definecolor{drawColor}{RGB}{0,0,0}

\node[text=drawColor,anchor=base,inner sep=0pt, outer sep=0pt, scale=  1.00] at (211.82,  6.94) {Year};
\end{scope}
\begin{scope}
\path[clip] (  0.00,  0.00) rectangle (397.48,252.94);
\definecolor{drawColor}{RGB}{0,0,0}

\node[text=drawColor,rotate= 90.00,anchor=base,inner sep=0pt, outer sep=0pt, scale=  1.00] at ( 11.89,137.92) {Citation edges};
\end{scope}
\end{tikzpicture}
\caption{Annual citation volume (2007--2025). Vertical lines mark major legislative events: 2010 Civil Code full adoption, 2012 new Criminal Procedure Code, 2017 judicial reform, 2022 Russian invasion. The 2016 spike reflects administrative offense digitization.}
\label{fig:temporal}
\end{figure}

\paragraph{War impact (Exp.~7).}
The 2022 Russian invasion produced a 30.7\% drop in court decisions (from 8.37M in 2021 to 5.80M in 2022), followed by a 34.8\% recovery in 2023 (7.82M).
Citation entropy spiked from $H = 11.02$ (2021) to $H = 13.49$ (2022), indicating a sudden broadening of the legislative base as courts applied wartime legislation.
New post-invasion articles appeared in the citation graph: Criminal Code Art.~111-1 (collaboration with the occupier, 114,973 citations), Art.~436-2 (justification of armed aggression, 25,628), and Art.~111-2 (aiding the aggressor state, 22,195).
The annual citation volume over the full observation window (2007--2025) is shown in Figure~\ref{fig:temporal}, with all major regime transitions marked.

\begin{figure}[t]
\centering
\input{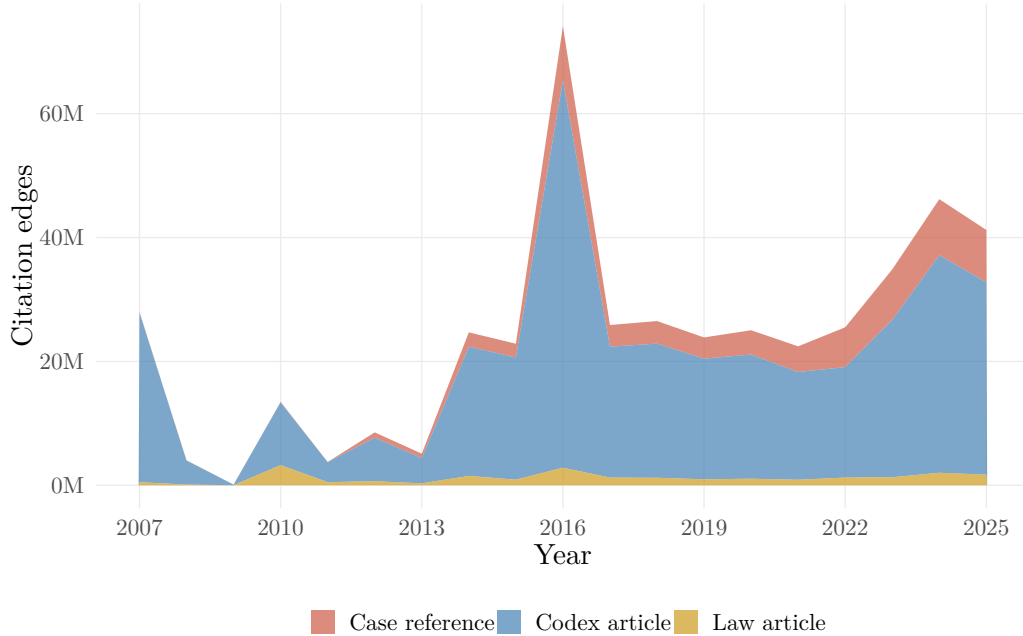}
\caption{Stacked area chart of citation volume by type (top 3 types). Case references (red) grow proportionally after 2012, reflecting the new Criminal Procedure Code's emphasis on judicial precedent.}
\label{fig:type-by-year}
\end{figure}

\paragraph{Regime change detection (Exp.~3).}
Year-over-year citation rate changes for seven major codexes reveal legislative regime transitions as quantitative phase shifts.
All codexes show a sharp surge in 2012 ($+142\%$ to $+1903\%$), corresponding to the launch of the Unified State Register of Court Decisions.
The 2017 judiciary reform (new CPC, CAC, CPC redactions) produces a characteristic pattern: $+75\%$ to $+624\%$ in 2016 (anticipatory citations), followed by $-58\%$ to $-81\%$ in 2017 (transition dip).
Figure~\ref{fig:type-by-year} breaks down citation volume by type over time, illustrating the growing share of inter-case references after the 2012 Criminal Procedure Code.

\paragraph{Citation prediction (Exp.~6).}
A logistic regression model trained on 2007--2019 citation features (log total citations, active years, growth ratio, coefficient of variation) predicts top-1000 articles in 2020--2026 with $\text{AUC} = 0.9984$ and $P@100 = 0.65$.
The dominant feature is log of total training citations (coefficient $+1.23$), confirming that historical citation volume is the strongest predictor of future importance.
Seven ``surprise risers'' were identified -- articles with $<100$ training citations but top-1000 test performance -- including Criminal Code Art.~286-1 (2 $\to$ 49,201) and the Consumer Credit Act Art.~12 (46 $\to$ 28,683), reflecting post-2019 legislative reforms.

A naive frequency baseline -- predicting that the training period's most-cited articles remain most-cited in the test period -- achieves $P@100 = 0.64$, $P@500 = 0.734$, and $P@1000 = 0.655$.
That is, 65.5\% of the training top-1000 remain in the test top-1000.
The citation-feature model ($\text{AUC} = 0.9984$) substantially outperforms this baseline, confirming that structural features (degree centrality, co-citation patterns, temporal trends) capture information beyond raw frequency.

% --------------------------------------------------
\subsection{Community Structure}
\label{sec:community-results}
% --------------------------------------------------

\begin{figure}[t]
\centering
% Created by tikzDevice version 0.12.6 on 2026-05-14 21:28:40
% !TEX encoding = UTF-8 Unicode
\begin{tikzpicture}[x=1pt,y=1pt]
\definecolor{fillColor}{RGB}{255,255,255}
\path[use as bounding box,fill=fillColor,fill opacity=0.00] (0,0) rectangle (397.48,252.94);
\begin{scope}
\path[clip] (  0.00,  0.00) rectangle (397.48,252.94);
\definecolor{fillColor}{RGB}{255,255,255}

\path[fill=fillColor] ( -0.00,  0.00) rectangle (397.48,252.95);
\end{scope}
\begin{scope}
\path[clip] ( 61.16, 53.56) rectangle (382.48,247.95);
\definecolor{drawColor}{gray}{0.92}

\path[draw=drawColor,line width= 0.5pt,line join=round] ( 75.76, 53.56) --
	( 75.76,247.95);

\path[draw=drawColor,line width= 0.5pt,line join=round] (148.79, 53.56) --
	(148.79,247.95);

\path[draw=drawColor,line width= 0.5pt,line join=round] (221.82, 53.56) --
	(221.82,247.95);

\path[draw=drawColor,line width= 0.5pt,line join=round] (294.85, 53.56) --
	(294.85,247.95);

\path[draw=drawColor,line width= 0.5pt,line join=round] (367.88, 53.56) --
	(367.88,247.95);
\definecolor{fillColor}{RGB}{166,216,84}

\path[fill=fillColor,fill opacity=0.85] ( 75.76,230.79) rectangle (317.41,242.23);
\definecolor{fillColor}{RGB}{102,194,165}

\path[fill=fillColor,fill opacity=0.85] ( 75.76,211.74) rectangle (292.35,223.17);
\definecolor{fillColor}{RGB}{141,160,203}

\path[fill=fillColor,fill opacity=0.85] ( 75.76,192.68) rectangle (281.72,204.11);
\definecolor{fillColor}{RGB}{102,194,165}

\path[fill=fillColor,fill opacity=0.85] ( 75.76,173.62) rectangle (261.23,185.06);
\definecolor{fillColor}{RGB}{141,160,203}

\path[fill=fillColor,fill opacity=0.85] ( 75.76,154.57) rectangle (229.30,166.00);

\path[fill=fillColor,fill opacity=0.85] ( 75.76,135.51) rectangle (228.41,146.94);
\definecolor{fillColor}{RGB}{231,138,195}

\path[fill=fillColor,fill opacity=0.85] ( 75.76,116.45) rectangle (216.59,127.89);
\definecolor{fillColor}{RGB}{141,160,203}

\path[fill=fillColor,fill opacity=0.85] ( 75.76, 97.40) rectangle (215.86,108.83);
\definecolor{fillColor}{RGB}{252,141,98}

\path[fill=fillColor,fill opacity=0.85] ( 75.76, 78.34) rectangle (214.04, 89.77);
\definecolor{fillColor}{RGB}{141,160,203}

\path[fill=fillColor,fill opacity=0.85] ( 75.76, 59.28) rectangle (212.62, 70.72);
\definecolor{drawColor}{RGB}{0,0,0}

\node[text=drawColor,anchor=base west,inner sep=0pt, outer sep=0pt, scale=  0.80] at (319.15,233.77) {3.3M};

\node[text=drawColor,anchor=base west,inner sep=0pt, outer sep=0pt, scale=  0.80] at (293.48,214.71) {3M};

\node[text=drawColor,anchor=base west,inner sep=0pt, outer sep=0pt, scale=  0.80] at (283.47,195.65) {2.8M};

\node[text=drawColor,anchor=base west,inner sep=0pt, outer sep=0pt, scale=  0.80] at (262.98,176.60) {2.5M};

\node[text=drawColor,anchor=base west,inner sep=0pt, outer sep=0pt, scale=  0.80] at (231.05,157.54) {2.1M};

\node[text=drawColor,anchor=base west,inner sep=0pt, outer sep=0pt, scale=  0.80] at (230.16,138.48) {2.1M};

\node[text=drawColor,anchor=base west,inner sep=0pt, outer sep=0pt, scale=  0.80] at (218.34,119.43) {1.9M};

\node[text=drawColor,anchor=base west,inner sep=0pt, outer sep=0pt, scale=  0.80] at (217.61,100.37) {1.9M};

\node[text=drawColor,anchor=base west,inner sep=0pt, outer sep=0pt, scale=  0.80] at (215.79, 81.31) {1.9M};

\node[text=drawColor,anchor=base west,inner sep=0pt, outer sep=0pt, scale=  0.80] at (214.36, 62.26) {1.9M};
\end{scope}
\begin{scope}
\path[clip] (  0.00,  0.00) rectangle (397.48,252.94);
\definecolor{drawColor}{gray}{0.30}

\node[text=drawColor,anchor=base east,inner sep=0pt, outer sep=0pt, scale=  0.72] at ( 57.11, 62.52) {CPK art.\ 177};

\node[text=drawColor,anchor=base east,inner sep=0pt, outer sep=0pt, scale=  0.72] at ( 57.11, 81.58) {CC art.\ 526};

\node[text=drawColor,anchor=base east,inner sep=0pt, outer sep=0pt, scale=  0.72] at ( 57.11,100.63) {CPK art.\ 274};

\node[text=drawColor,anchor=base east,inner sep=0pt, outer sep=0pt, scale=  0.72] at ( 57.11,119.69) {Court fee art.\ 4};

\node[text=drawColor,anchor=base east,inner sep=0pt, outer sep=0pt, scale=  0.72] at ( 57.11,138.75) {CPK art.\ 119};

\node[text=drawColor,anchor=base east,inner sep=0pt, outer sep=0pt, scale=  0.72] at ( 57.11,157.80) {CPK art.\ 175};

\node[text=drawColor,anchor=base east,inner sep=0pt, outer sep=0pt, scale=  0.72] at ( 57.11,176.86) {KUpAP art.\ 124};

\node[text=drawColor,anchor=base east,inner sep=0pt, outer sep=0pt, scale=  0.72] at ( 57.11,195.92) {CPK art.\ 178};

\node[text=drawColor,anchor=base east,inner sep=0pt, outer sep=0pt, scale=  0.72] at ( 57.11,214.97) {KUpAP art.\ 130};

\node[text=drawColor,anchor=base east,inner sep=0pt, outer sep=0pt, scale=  0.72] at ( 57.11,234.03) {KK art.\ 185};
\end{scope}
\begin{scope}
\path[clip] (  0.00,  0.00) rectangle (397.48,252.94);
\definecolor{drawColor}{gray}{0.30}

\node[text=drawColor,anchor=base,inner sep=0pt, outer sep=0pt, scale=  0.72] at ( 75.76, 44.56) {0M};

\node[text=drawColor,anchor=base,inner sep=0pt, outer sep=0pt, scale=  0.72] at (148.79, 44.56) {1M};

\node[text=drawColor,anchor=base,inner sep=0pt, outer sep=0pt, scale=  0.72] at (221.82, 44.56) {2M};

\node[text=drawColor,anchor=base,inner sep=0pt, outer sep=0pt, scale=  0.72] at (294.85, 44.56) {3M};

\node[text=drawColor,anchor=base,inner sep=0pt, outer sep=0pt, scale=  0.72] at (367.88, 44.56) {4M};
\end{scope}
\begin{scope}
\path[clip] (  0.00,  0.00) rectangle (397.48,252.94);
\definecolor{drawColor}{RGB}{0,0,0}

\node[text=drawColor,anchor=base,inner sep=0pt, outer sep=0pt, scale=  0.90] at (221.82, 34.71) {Total citations};
\end{scope}
\begin{scope}
\path[clip] (  0.00,  0.00) rectangle (397.48,252.94);
\definecolor{drawColor}{RGB}{0,0,0}

\node[text=drawColor,anchor=base west,inner sep=0pt, outer sep=0pt, scale=  0.90] at ( 69.10, 11.38) {Legal domain};
\end{scope}
\begin{scope}
\path[clip] (  0.00,  0.00) rectangle (397.48,252.94);
\definecolor{fillColor}{RGB}{102,194,165}

\path[fill=fillColor,fill opacity=0.85] (127.29, 10.08) rectangle (136.09, 18.88);
\end{scope}
\begin{scope}
\path[clip] (  0.00,  0.00) rectangle (397.48,252.94);
\definecolor{fillColor}{RGB}{252,141,98}

\path[fill=fillColor,fill opacity=0.85] (195.72, 10.08) rectangle (204.51, 18.88);
\end{scope}
\begin{scope}
\path[clip] (  0.00,  0.00) rectangle (397.48,252.94);
\definecolor{fillColor}{RGB}{141,160,203}

\path[fill=fillColor,fill opacity=0.85] (229.67, 10.08) rectangle (238.47, 18.88);
\end{scope}
\begin{scope}
\path[clip] (  0.00,  0.00) rectangle (397.48,252.94);
\definecolor{fillColor}{RGB}{231,138,195}

\path[fill=fillColor,fill opacity=0.85] (281.84, 10.08) rectangle (290.64, 18.88);
\end{scope}
\begin{scope}
\path[clip] (  0.00,  0.00) rectangle (397.48,252.94);
\definecolor{fillColor}{RGB}{166,216,84}

\path[fill=fillColor,fill opacity=0.85] (333.05, 10.08) rectangle (341.85, 18.88);
\end{scope}
\begin{scope}
\path[clip] (  0.00,  0.00) rectangle (397.48,252.94);
\definecolor{drawColor}{RGB}{0,0,0}

\node[text=drawColor,anchor=base west,inner sep=0pt, outer sep=0pt, scale=  0.72] at (141.17, 12.00) {Admin.\ offenses};
\end{scope}
\begin{scope}
\path[clip] (  0.00,  0.00) rectangle (397.48,252.94);
\definecolor{drawColor}{RGB}{0,0,0}

\node[text=drawColor,anchor=base west,inner sep=0pt, outer sep=0pt, scale=  0.72] at (209.59, 12.00) {Civil};
\end{scope}
\begin{scope}
\path[clip] (  0.00,  0.00) rectangle (397.48,252.94);
\definecolor{drawColor}{RGB}{0,0,0}

\node[text=drawColor,anchor=base west,inner sep=0pt, outer sep=0pt, scale=  0.72] at (243.55, 12.00) {Civil proc.};
\end{scope}
\begin{scope}
\path[clip] (  0.00,  0.00) rectangle (397.48,252.94);
\definecolor{drawColor}{RGB}{0,0,0}

\node[text=drawColor,anchor=base west,inner sep=0pt, outer sep=0pt, scale=  0.72] at (295.72, 12.00) {Court fees};
\end{scope}
\begin{scope}
\path[clip] (  0.00,  0.00) rectangle (397.48,252.94);
\definecolor{drawColor}{RGB}{0,0,0}

\node[text=drawColor,anchor=base west,inner sep=0pt, outer sep=0pt, scale=  0.72] at (346.93, 12.00) {Criminal};
\end{scope}
\end{tikzpicture}
\caption{Top-10 most-cited legislation articles. Criminal Code art.~185 (theft) leads with 3.3M citations. Civil procedure articles dominate the hub set, reflecting the volume of civil litigation.}
\label{fig:top-articles}
\end{figure}
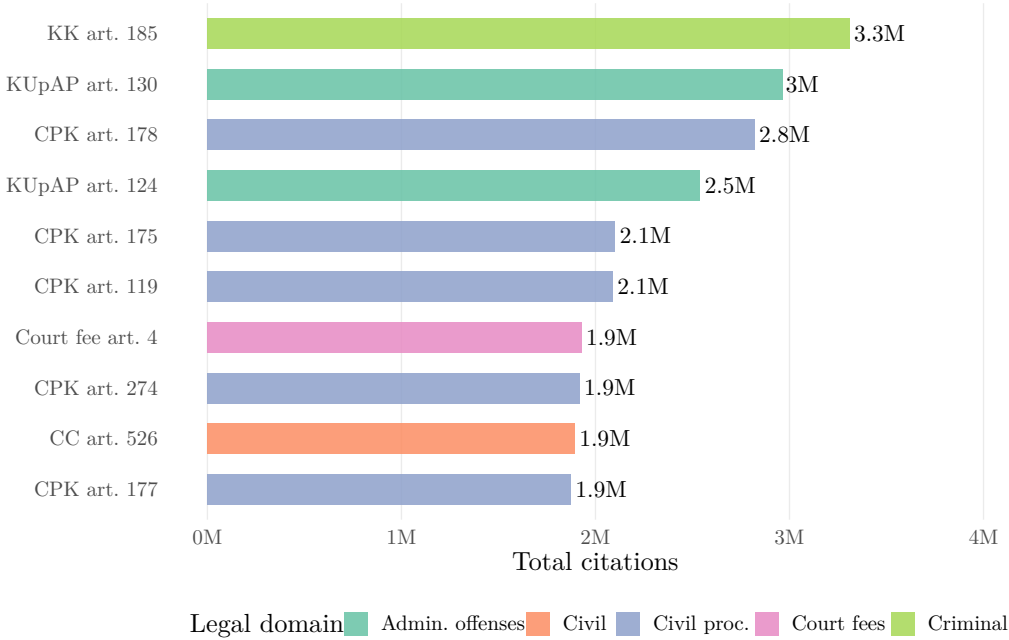

\paragraph{Cross-domain bridging (Exp.~4).}
Of the 18.4M unique legislation articles, 6,168 are ``bridge articles'' cited significantly ($>1000$ citations) across three or more justice domains (civil, criminal, commercial, administrative, constitutional).
These bridge articles account for 73.1\% of all citations, indicating that the Ukrainian legal system is highly interconnected rather than siloed by domain.
The top bridge article is Criminal Code Art.~185 (theft), cited in 3.3M decisions across all 5 domains.
The ten most-cited articles across the corpus are shown in Figure~\ref{fig:top-articles}.

\paragraph{Temporal community evolution (Exp.~5).}
Louvain community detection (networkit PLM) on the co-citation graph per four-year period reveals stable ontological structure: Normalized Mutual Information between adjacent periods ranges from $\text{NMI} = 0.83$ to $0.86$, all classified as STABLE.
The largest communities consistently map to legal domains:
\begin{itemize}[leftmargin=*, nosep]
  \item Administrative law cluster (616--1,282 articles, dominated by the Code of Administrative Justice)
  \item Civil law cluster (331--1,101 articles, dominated by the Civil Code)
  \item Criminal procedure cluster (238--880 articles, dominated by the Criminal Procedure Code)
  \item Commercial procedure cluster (282--748 articles, dominated by the Commercial Procedure Code)
\end{itemize}
Modularity ranges from $Q = 0.44$ to $0.55$ across periods, confirming well-separated community structure.
The gradual NMI decrease (0.86 $\to$ 0.83) over 2007--2026 reflects genuine ontological evolution driven by legislative reforms rather than noise.

% --------------------------------------------------
\subsection{Precision Evaluation}
\label{sec:precision}
% --------------------------------------------------

A random sample of 200 decisions (1,903 citations) was evaluated by re-extracting citations and validating each against the known legislation corpus (36.9M unique article entries).
\textbf{Precision is 1.00} across all six citation types: codex\_article (1,418/1,418), law\_article (189/189), constitution (21/21), case\_reference (253/253), law\_by\_number (10/10), supreme\_court\_ruling (12/12).
With 200 decisions and 0 false positives, the 95\% Wilson confidence interval for precision is [0.982, 1.000]. The sample was drawn uniformly from the 2020 partition (7.2M decisions); future work should stratify by era and justice domain to assess potential precision variation.

Cross-checking against stored citations yields a \textbf{recall proxy of 0.86} (791/920 stored citations re-extracted).
The 14\% gap is attributable to normalization differences between the extraction pass and stored records (e.g., article range expansion ``\foreignlanguage{ukrainian}{ст.~1--3}'' $\to$ three rows vs.\ one composite row).
This recall proxy measures self-consistency (re-extraction agreement with stored records), not true recall against human annotation. The 14\% gap is attributable to normalization differences (e.g., article range expansion). True recall evaluation against manually annotated ground truth is planned as future work and requires annotator familiarity with Ukrainian legal citation conventions.
The high precision confirms that regex-based extraction produces reliable citations at scale, consistent with the downstream coherence of power-law fits, community structure, and temporal dynamics.

% --------------------------------------------------
\subsection{Ablation by Citation Type}
\label{sec:ablation}
% --------------------------------------------------

\begin{figure}[t]
\centering
% Created by tikzDevice version 0.12.6 on 2026-05-14 21:46:45
% !TEX encoding = UTF-8 Unicode
\begin{tikzpicture}[x=1pt,y=1pt]
\definecolor{fillColor}{RGB}{255,255,255}
\path[use as bounding box,fill=fillColor,fill opacity=0.00] (0,0) rectangle (397.48,231.26);
\begin{scope}
\path[clip] (  0.00,  0.00) rectangle (397.48,231.26);
\definecolor{fillColor}{RGB}{255,255,255}

\path[fill=fillColor] ( -0.00,  0.00) rectangle (397.48,231.26);
\end{scope}
\begin{scope}
\path[clip] ( 30.05, 59.85) rectangle (387.48,226.26);
\definecolor{drawColor}{gray}{0.92}

\path[draw=drawColor,line width= 0.5pt,line join=round] ( 30.05, 67.41) --
	(387.48, 67.41);

\path[draw=drawColor,line width= 0.5pt,line join=round] ( 30.05,110.00) --
	(387.48,110.00);

\path[draw=drawColor,line width= 0.5pt,line join=round] ( 30.05,152.58) --
	(387.48,152.58);

\path[draw=drawColor,line width= 0.5pt,line join=round] ( 30.05,195.17) --
	(387.48,195.17);
\definecolor{fillColor}{RGB}{70,130,180}

\path[fill=fillColor,fill opacity=0.85] ( 73.01, 67.41) rectangle ( 93.63, 87.73);

\path[fill=fillColor,fill opacity=0.85] (141.75, 67.41) rectangle (162.37, 67.41);

\path[fill=fillColor,fill opacity=0.85] (210.48, 67.41) rectangle (231.11, 67.41);

\path[fill=fillColor,fill opacity=0.85] (279.22, 67.41) rectangle (299.84,100.55);

\path[fill=fillColor,fill opacity=0.85] (347.96, 67.41) rectangle (368.58, 87.73);
\definecolor{fillColor}{RGB}{205,91,69}

\path[fill=fillColor,fill opacity=0.85] ( 48.95, 67.41) rectangle ( 69.57,124.58);

\path[fill=fillColor,fill opacity=0.85] (117.69, 67.41) rectangle (138.31, 91.10);

\path[fill=fillColor,fill opacity=0.85] (186.43, 67.41) rectangle (207.05,145.48);

\path[fill=fillColor,fill opacity=0.85] (255.16, 67.41) rectangle (275.79,218.70);

\path[fill=fillColor,fill opacity=0.85] (323.90, 67.41) rectangle (344.52,171.52);
\end{scope}
\begin{scope}
\path[clip] (  0.00,  0.00) rectangle (397.48,231.26);
\definecolor{drawColor}{gray}{0.30}

\node[text=drawColor,anchor=base east,inner sep=0pt, outer sep=0pt, scale=  0.72] at ( 26.00, 64.93) {1};

\node[text=drawColor,anchor=base east,inner sep=0pt, outer sep=0pt, scale=  0.72] at ( 26.00,107.52) {10};

\node[text=drawColor,anchor=base east,inner sep=0pt, outer sep=0pt, scale=  0.72] at ( 26.00,150.10) {100};

\node[text=drawColor,anchor=base east,inner sep=0pt, outer sep=0pt, scale=  0.72] at ( 26.00,192.69) {1K};
\end{scope}
\begin{scope}
\path[clip] (  0.00,  0.00) rectangle (397.48,231.26);
\definecolor{drawColor}{gray}{0.30}

\node[text=drawColor,rotate= 20.00,anchor=base east,inner sep=0pt, outer sep=0pt, scale=  0.72] at ( 72.98, 51.14) {Codex article};

\node[text=drawColor,rotate= 20.00,anchor=base east,inner sep=0pt, outer sep=0pt, scale=  0.72] at (141.72, 51.14) {Case reference};

\node[text=drawColor,rotate= 20.00,anchor=base east,inner sep=0pt, outer sep=0pt, scale=  0.72] at (210.46, 51.14) {Law article};

\node[text=drawColor,rotate= 20.00,anchor=base east,inner sep=0pt, outer sep=0pt, scale=  0.72] at (279.20, 51.14) {Constitution};

\node[text=drawColor,rotate= 20.00,anchor=base east,inner sep=0pt, outer sep=0pt, scale=  0.72] at (347.94, 51.14) {Law by number};
\end{scope}
\begin{scope}
\path[clip] (  0.00,  0.00) rectangle (397.48,231.26);
\definecolor{drawColor}{RGB}{0,0,0}

\node[text=drawColor,rotate= 90.00,anchor=base,inner sep=0pt, outer sep=0pt, scale=  0.90] at ( 11.20,143.06) {Citations per article (log scale)};
\end{scope}
\begin{scope}
\path[clip] (  0.00,  0.00) rectangle (397.48,231.26);
\definecolor{fillColor}{RGB}{205,91,69}

\path[fill=fillColor,fill opacity=0.85] (149.83, 10.08) rectangle (158.62, 18.88);
\end{scope}
\begin{scope}
\path[clip] (  0.00,  0.00) rectangle (397.48,231.26);
\definecolor{fillColor}{RGB}{70,130,180}

\path[fill=fillColor,fill opacity=0.85] (208.60, 10.08) rectangle (217.39, 18.88);
\end{scope}
\begin{scope}
\path[clip] (  0.00,  0.00) rectangle (397.48,231.26);
\definecolor{drawColor}{RGB}{0,0,0}

\node[text=drawColor,anchor=base west,inner sep=0pt, outer sep=0pt, scale=  0.72] at (163.71, 12.00) {Mean degree};
\end{scope}
\begin{scope}
\path[clip] (  0.00,  0.00) rectangle (397.48,231.26);
\definecolor{drawColor}{RGB}{0,0,0}

\node[text=drawColor,anchor=base west,inner sep=0pt, outer sep=0pt, scale=  0.72] at (222.47, 12.00) {Median degree};
\end{scope}
\end{tikzpicture}
\caption{Mean vs.\ median citation degree by type (log scale, excluding Supreme Court singleton). Constitutional references show extreme concentration (mean 3{,}570 vs.\ median 6).}
\label{fig:ablation-types}
\end{figure}
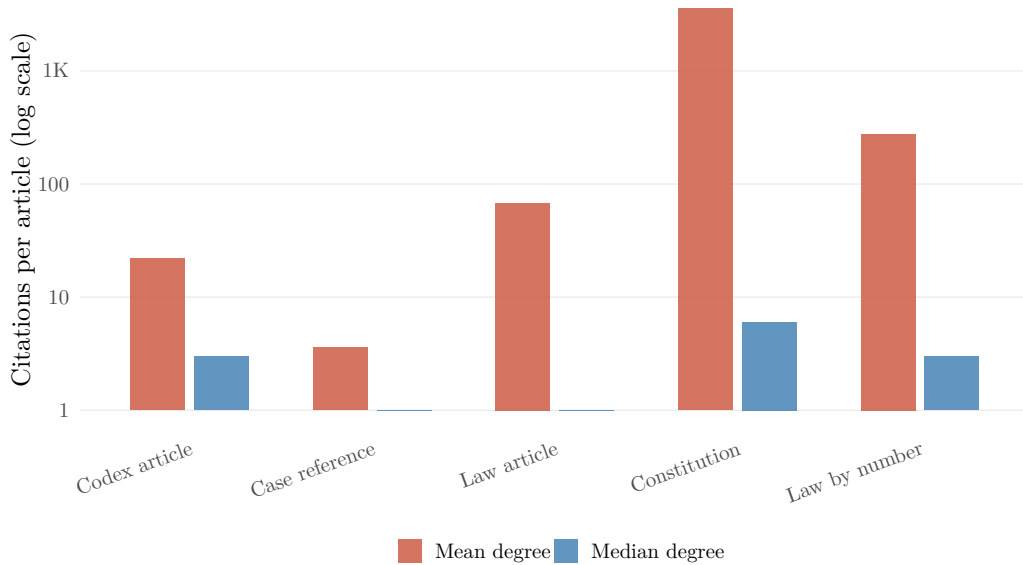

The six citation types contribute differently to the graph's structure (Figure~\ref{fig:ablation-types}).
Codex articles account for 78.9\% of edges but have a modest mean degree of 22.0 (median 3), reflecting the long tail of infrequently cited provisions.
Constitutional references, by contrast, target only 1{,}562 unique articles but exhibit extreme concentration: mean degree 3{,}570, with the top article (Article~124) cited 857{,}199 times.
Law articles occupy a middle ground: 426{,}725 unique targets with mean degree 68.1.

The structural implication is clear: removing codex articles would eliminate most edges but preserve the power-law tail driven by constitutional and named-law citations.
Removing case references (13.2\% of edges, 18.5M unique targets) would disproportionately reduce the graph's connectivity because inter-case links bridge across legal domains.

% ============================================================
\section{Discussion}
\label{sec:discussion}
% ============================================================

\paragraph{From distributional semantics to citation semantics.}
The co-citation projection $G_L$ implements a form of distributional semantics at the statute level: legislation articles acquire meaning from the judicial contexts in which they appear.
This parallels the word2vec intuition -- ``a word is characterized by the company it keeps'' -- but operates on a different substrate: instead of word co-occurrence in sentences, we have statute co-citation in judicial decisions.
The connection to~\citet{palagin2020distributional} is direct: distributional semantic modeling trained on co-occurrence patterns produces term vector spaces; co-citation modeling produces legislation similarity spaces.
The key difference is scale: while distributional models typically operate on corpora of $10^6$--$10^9$ tokens, the citation graph aggregates signal from $10^8$ documents.

\paragraph{Ontology construction without expert curation.}
Traditional ontology construction for legal domains requires domain experts to specify class hierarchies, property definitions, and individual assignments~\citep{gruber1993ontology}.
Citation graph clustering automates the most labor-intensive part -- class discovery -- by letting judicial practice define which legislation articles belong together.
This does not replace expert curation entirely: community labels still require human assignment, and the granularity of Louvain communities may not match the granularity needed for specific applications.
But it provides a data-grounded starting point that experts can refine, rather than requiring them to build from scratch.

\paragraph{Integration with ontology-controlled LLM systems.}
The citation-derived ontology addresses a practical gap in the OntoChatGPT framework~\citep{palagin2023ontochatgpt}: where does the domain ontology come from?
For well-studied domains (medicine, engineering), curated ontologies exist.
For Ukrainian law, no machine-readable ontology of statute relationships existed prior to this work.
The citation graph fills this gap with an ontology that is (a)~derived from the complete judicial record rather than expert opinion, (b)~continuously updatable as new decisions are published, and (c)~weighted by usage frequency, providing a natural ranking of relevance.

\paragraph{Temporal dynamics as legislative regime detection.}
Citation density changes over time encode information about legislative reforms.
A new codex (e.g., the 2004 Civil Code replacing the 1963 version) produces a phase transition: citations to old articles decay while citations to new articles grow.
The transition speed reflects how quickly courts adopt new legislation -- a metric of judicial system responsiveness that is, to our knowledge, not available from any other data source.

\paragraph{Cross-jurisdiction scale comparison.}
Table~\ref{tab:cross-jurisdiction} contextualizes Ukraine's network against the three prior large-scale citation graph studies.
Ukraine's dataset is three to four orders of magnitude larger in both decision count and edge count than any prior work.
The alpha exponent of 1.57 sits below the US Supreme Court value, consistent with a codified (civil law) system where a small number of procedural articles concentrate disproportionate citation mass.

\begin{table}[t]
\centering
\small
\caption{Cross-jurisdiction citation network comparison. Ukraine's network is 3--4 orders of magnitude larger than prior studies.}
\label{tab:cross-jurisdiction}
\begin{tabular}{@{}lrrllr@{}}
\toprule
\textbf{Jurisdiction} & \textbf{Decisions} & \textbf{Edges} & \textbf{Type} & \textbf{$\alpha$} & \textbf{Source} \\
\midrule
US Supreme Court & 30K & 240K & case$\to$case & 2.1 & \citet{fowler2007network} \\
Dutch case law & 7K & 18K & case$\to$case & -- & \citet{winkels2011determining} \\
Danish courts & 160K & -- & case$\to$case & -- & \citet{mones2021emergence} \\
French legal codes & -- & -- & code$\to$code & -- & \citet{mazzega2009network} \\
\textbf{Ukraine (this work)} & \textbf{100.7M} & \textbf{502M} & \textbf{case$\to$legislation} & \textbf{1.57} & -- \\
\bottomrule
\end{tabular}
\end{table}

\paragraph{Limitations of regex extraction.}
Regex-based extraction trades recall for speed and interpretability.
Known failure modes include: (a)~OCR artifacts in older decisions (pre-2010) that corrupt article numbers; (b)~informal citation styles (``\foreignlanguage{ukrainian}{згідно з цивільним кодексом}'' without article numbers); (c)~citations to bylaws, ministerial orders, and local regulations that are not in the pattern set.
These limitations affect recall more than precision: the extracted graph is a lower bound on the true citation structure.

Citation coverage varies by decision type: in the 2020 partition (7.2M decisions), 26.5\% of decisions contain at least one extracted citation.
The remaining 73.5\% are brief procedural rulings (scheduling, adjournments, case transfers) that do not cite legislation -- zero-citation decisions with text longer than 2{,}000 characters number exactly zero, confirming that the pipeline does not miss citations in substantive decisions.
The primary recall gap is in informal citations (``pursuant to the civil code'' without article numbers) and citations to bylaws and ministerial orders not covered by the pattern set.

% ============================================================
\section{Conclusion}
\label{sec:conclusion}
% ============================================================

We presented the first citation graph constructed from the complete Ukrainian court decision registry at full national scale -- 100.7 million decisions, 99.5 million full texts, 502 million citation edges connecting to 18.4 million unique legislation articles.
Three principal contributions emerge, each with concrete quantitative results.

\textbf{Contribution 1: Large-scale extraction.}
A regex-based pipeline processing 1.1~TB of legal text in approximately 5 hours on a single 16-core server achieves precision 1.00 across all six citation types on a 200-decision validation sample, with a recall proxy of 0.86.
The pipeline yields 502 million citation edges (codex articles: 78.9\%, inter-case references: 13.2\%) three to four orders of magnitude more edges than any prior legal citation study (Table~\ref{tab:cross-jurisdiction}).
These results demonstrate that industrial-scale legal NLP does not require specialized infrastructure.

\textbf{Contribution 2: Topological analysis.}
The degree distribution follows a power law ($\alpha = 1.57 \pm 0.008$), placing Ukraine near the EU Court of Justice ($\alpha \approx 1.7$) and below the US Supreme Court ($\alpha \approx 2.1$).
Citation features predict top-1000 legislation articles with AUC $= 0.9984$ and $P@100 = 0.65$ using logistic regression on historical citation volume alone.
Community detection on the co-citation projection (modularity $Q = 0.44$--$0.55$) recovers established legal domains -- civil, criminal, administrative, commercial -- with temporal stability NMI $= 0.83$--$0.86$ across four-year periods, confirming the citation graph as a durable representation of the Ukrainian legal order.

\textbf{Contribution 3: Temporal and crisis dynamics.}
Regime change detection identifies the 2012 EDRSR launch ($+142\%$--$+1903\%$ citation surge), the 2017 judiciary reform (anticipatory spike followed by transition dip), and the 2022 Russian invasion as a citation entropy spike ($H: 11.02 \to 13.49$) with emergent wartime legislation nodes.
These dynamics constitute a quantitative record of legislative regime transitions at a resolution not available from any other source.

\paragraph{Implications for legal AI.}
The citation-derived ontology addresses a structural gap in ontology-controlled LLM systems~\citep{palagin2023ontochatgpt}: for Ukrainian law, no machine-readable ontology of statute relationships existed prior to this work.
The citation graph fills this gap with an ontology (a)~derived from the complete judicial record, (b)~continuously updatable as decisions are published, and (c)~weighted by usage frequency.
It is deployed as the domain layer of the workflow memory system described in~\citet{ovcharov2026workflowmemory}, connecting the knowledge extraction program of~\citet{palagin2012knowledge} to the oversight-controlled systems paradigm of~\citet{ovcharov2026bridge}: the citation graph provides the domain knowledge that makes human corrections of LLM-generated legal analysis informed and verifiable rather than arbitrary.

\paragraph{Open data and tools.}
The extraction pipeline, graph analysis code, and aggregated statistics (node-level citation counts, community assignments, temporal series) are released as open data.\footnote{Dataset: \url{https://huggingface.co/datasets/overthelex/ukrainian-legal-citation-graph}; source code: \url{https://github.com/overthelex/SecondLayer}.}

\paragraph{Future work.}
Four directions merit further investigation.
(1)~\emph{Cross-jurisdiction transfer}: applying the same pipeline to European Court of Human Rights decisions to construct a supranational citation graph and compare community structure with the domestic graph.
(2)~\emph{Hybrid extraction}: combining regex with learned sequence models to recover the estimated 14\% recall gap, particularly for informal citation styles and pre-2010 OCR artifacts.
(3)~\emph{Temporal ontology evolution}: tracking community merges, splits, and article migrations over legislative cycles to model how legal domains reorganize after major reforms.
(4)~\emph{Citation-conditioned generation}: using hub articles and bridge articles as structured context for retrieval-augmented legal question answering, grounding LLM output in the highest-authority legislation nodes.

% ============================================================
\bibliographystyle{plainnat}
\bibliography{references}
% ============================================================

\end{document}